\documentclass[fleqn]{article}
\usepackage{spconf,graphicx}
\usepackage[fleqn]{amsmath}
\usepackage{amsfonts}
\usepackage{amssymb}
\usepackage{url}
\usepackage{float}
\usepackage{dsfont}

\newcommand{\diff}{\mathop{}\!d}

\title{SEQUENCE-TO-SEQUENCE ASR OPTIMIZATION VIA REINFORCEMENT LEARNING}
\name{Andros Tjandra\textsuperscript{1}, Sakriani Sakti\textsuperscript{1,2}, Satoshi Nakamura\textsuperscript{1,2}}
\address{\textsuperscript{1} Graduate School of Information Science, Nara Institute of Science and Technology, Japan\\\textsuperscript{2} RIKEN, Center for Advanced Intelligence Project AIP, Japan\\
\texttt{\{andros.tjandra.ai6, ssakti, s-nakamura\}@is.naist.jp}}
\begin{document}
\ninept
\maketitle
\begin{abstract}
Despite the success of sequence-to-sequence approaches in automatic speech recognition (ASR) systems, the models still suffer from several problems, mainly due to the mismatch between the training and inference conditions. In the sequence-to-sequence architecture, the model is trained to predict the grapheme of the current time-step given the input of speech signal and the ground-truth grapheme history of the previous time-steps. However, it remains unclear how well the model approximates real-world speech during inference. Thus, generating the whole transcription from scratch based on previous predictions is complicated and errors can propagate over time. Furthermore, the model is optimized to maximize the likelihood of training data instead of error rate evaluation metrics that actually quantify recognition quality. This paper presents an alternative strategy for training sequence-to-sequence ASR models by adopting the idea of reinforcement learning (RL). Unlike the standard training scheme with maximum likelihood estimation, our proposed approach utilizes the policy gradient algorithm. We can (1) sample the whole transcription based on the model's prediction in the training process and (2) directly optimize the model with negative Levenshtein distance as the reward. Experimental results demonstrate that we significantly improved the performance compared to a model trained only with maximum likelihood estimation.
\end{abstract}
\section{Introduction}
\vspace{-0.2cm}
\label{sec:intro}
Sequence-to-sequence models have been recently shown to be very effective for many tasks such as machine translation \cite{sutskever2014sequence, bahdanau2014neural}, image captioning \cite{xu2015show, vinyals2015show}, and speech recognition \cite{bahdanau2016end}. With these models, we are able to learn a direct mapping between the variable-length of the source and the target sequences that are often not known apriori using only a single neural network architecture. This way, many complicated hand-engineered models can also be simplified by letting DNNs find their way to map from input to output spaces \cite{bahdanau2016end, chan2016listen, tjandra2017attention}. Therefore, we can eliminate the need to construct separate components, i.e., a feature extractor, an acoustic model, a lexicon model, or a language model, as is commonly required in conventional ASR systems such as hidden Markov model-Gaussian mixture model (HMM-GMM)-based or hybrid HMM-DNN.

A generic sequence-to-sequence model commonly consists of three modules: (1) an encoder module for representing source data information, (2) a decoder module for generating transcription output and (3) an attention module for extracting related information from an encoder representation based on the current decoder state. A decoding scheme was done based on a left-to-right decoding procedure. In the training stage, given the current input of the speech signal, the decoder produces a grapheme in the current time-step with maximal probability conditioned on the ground-truth of the grapheme history in the previous time-steps. This training scheme is usually referred as a teacher-forcing method \cite{williams1989learning}. However, in the inference stage, since the ground-truth of the transcription is not known, the model must produce the grapheme in the current time-step based on an approximation of the correct grapheme in previous time-steps. Therefore, an incorrect decision in an earlier time-step may propagate through subsequent time-steps.

Another drawback is the differences in the use of objective functions between training and evaluation schemes. In the training stage, the model is mostly optimized by combining the teacher-forcing approach with the maximum likelihood estimation (MLE) for each frame. On the other hand, the recognition accuracy is evaluated by calculating the minimum string edit-distance (Levenshtein distance) between the correct transcription and the recognition output. Such differences may result in suboptimal performance \cite{wu2016google}. Optimizing the model parameter with the appropriate objective function is crucial to achieve good model performance, or in other words, direct optimization with respect to the evaluation metrics might be necessary.

In this paper, we propose an alternative strategy for training a sequence-to-sequence ASR by adopting an idea from RL. Specifically, we utilize a policy gradient algorithm (REINFORCE) \cite{williams1992simple} to simultaneously alleviate both of the above problems. By treating our decoder as a policy network or an agent, we are able to (1) sample the whole transcription based on model's prediction in the training process and (2) directly optimize the model with negative Levenshtein distance as the reward. Our model thus integrates the power of the sequence-to-sequence approach to learn the mapping between the speech signal and the text transcription based on the strength of reinforcement learning to optimize the model with ASR performance metric directly.

\section{Sequence-to-Sequence ASR}
\vspace{-0.2cm}
\label{sec:encdecasr}

\begin{figure}[]
	\centering
	\includegraphics[width=0.70\linewidth]{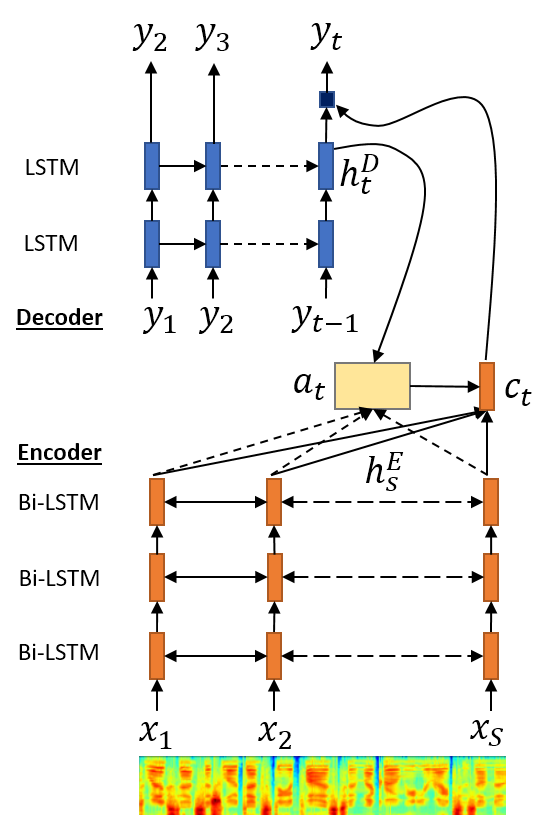}
	\caption{Attention-based encoder-decoder architecture.}
	\label{fig:atte2e}
	\vspace{-0.3cm}
\end{figure}
Sequence-to-sequence model is a type of neural network model that directly models conditional probability $P(\mathbf{y}|\mathbf{x})$, where $\mathbf{x} = [x_1, ..., x_S]$ is the source sequence with length $S$, and $\mathbf{y} = [y_1, ..., y_T]$ is the target sequence with length $T$. Most common input $\mathbf{x}$ is a sequence of feature vectors like Mel-spectral filterbank and/or MFCC. Therefore, $\mathbf{x} \in \mathbb{R}^{S \times F}$ where F is the number of features and S is the total frame length for an utterance. Output $\mathbf{y}$, which is a speech transcription sequence, can be either a phoneme or a grapheme (character) sequence.

Figure~\ref{fig:atte2e} shows the overall structure of the attention-based encoder-decoder model that consists of encoder, decoder, and attention modules. The encoder task processes input sequence $\mathbf{x}$ and outputs representative information $\mathbf{h^E} = [h^E_1, ...,h^E_S]$ for the decoder. The attention module is an extension scheme that helps the decoder find relevant information on the encoder side based on current decoder hidden states \cite{bahdanau2014neural}. An attention module produces context information $c_t$ at time $t$ based on the encoder and decoder hidden states with following equation:
\begin{align}
c_t &= \sum_{s=1}^{S} a_t(s) * h^E_s \\
a_t(s) &= \text{Align}({h^E_s}, h^D_t) = \frac{\exp(\text{Score}(h^E_s, h^D_t))}{\sum_{s=1}^{S}\exp(\text{Score}(h^E_s, h^D_t))} \label{eq:align}.
\end{align}
There are several variations for the score functions:
\begin{align}
\text{Score}(h_s^E, h_t^D) =
\begin{cases}
\langle h_s^E, h_t^D\rangle, & \text{dot product}  \\
h_s^{E\intercal} W_{s} h_t^D, & \text{bilinear}  \\
V_s^{\intercal} \tanh(W_{s} [h_s^E, h_t^D]), & \text{MLP} \label{eq:mlpscore} \\
\end{cases}
\end{align} where $\text{Score}:(\mathbb{R}^M \times \mathbb{R}^N) \rightarrow \mathbb{R}$, $M$ is the number of hidden units for the encoder and $N$ is the number of hidden units for the decoder.
Finally, the decoder task, which predicts the target sequence probability at time $t$ based on the previous output and context information $c_t$ can be formulated:
\begin{equation}
\log{P(\mathbf{y}|\mathbf{x}; \theta)} = \sum_{t=1}^{T}\log{P(y_t|h_t^D, c_t; \theta)}
\end{equation} where $h_t^D$ is the last decoder layer that contains summarized information from all previous input $\mathbf{y}_{<t}$ and $\theta$ is our model parameters.

\section{Sequence-to-Sequence Optimization with Reinforcement Learning}
\vspace{-0.2cm}
In this section, we introduce our proposed approach that integrates policy optimization with the standard encoder-decoder ASR model. We start by describing the policy gradient method and followed by the reward construction for our ASR agent.

\subsection{Policy Gradient}
\vspace{-0.2cm}
Policy gradient is a type of reinforcement learning algorithm for optimizing the expected rewards with respect to the parameterized policy \cite{suttonrlbook}. To apply the idea from the policy gradient method, we need to establish a connection between our ASR model and the reinforcement learning formulation. For reinforcement learning, we reformulate our system as a Markov Decision Process (MDP) $= (\mathcal{S}, \mathcal{A}, \mathcal{T}, \mathcal{R})$, where $\mathcal{S}$ is the state space, $\mathcal{A}$ is the set of possible actions, $\mathcal{T}$ is the transition probability, and $\mathcal{R}$ is the reward function.

Here, our task is to generate a text transcription given the input speech waveform, and the encoder-decoder neural network (Section ~\ref{sec:encdecasr}) will act as an agent. For each time-step $t=1,2,3...,T$, we can define state $s_t \in \mathcal{S}$ as $s_t = [h_t^D, c_t]$, which is the concatenation between the decoder hidden state and the context information at time $t$. Action $a_t \in \mathcal{A}$ equals $a_t = y_t$, where action space $\mathcal{A}$ contains all possible grapheme + end of sentence ``eos'' symbols in our dataset. Reward function $\mathcal{R}$ for our ASR task will be explained later in Section~\ref{sec:reward}.

Given a pair of speech and transcription $(\mathbf{x}^{(n)}, \mathbf{y}^{(n)})$ at $n$-th index, $R^{(n)}$ is the reward for transcription $\mathbf{y}$ compared to ground-truth $\mathbf{y}^{(n)}$. Our optimization target is to maximize expected reward $E_{\mathbf{y}}[R^{(n)}|\pi_\theta]$ with respect to $\theta$ as our neural network parameter where $\pi_\theta(a_t|s_t) = P(y_t|h_t^{D(n)}, c_t^{(n)}; \theta) = P(y_t|
\mathbf{y}_{<t}, \mathbf{x}^{(n)}; \theta)$. To use the first-order optimization method (e.g., stochastic gradient ascent / descent), we need to calculate the gradient from the expected rewards:
\begin{align}
&\nabla_\theta E_{\mathbf{y}}\left[ R^{(n)} | \pi_\theta \right] = \nabla_\theta \int P(\mathbf{y}|\mathbf{x}^{(n)}; \theta) R^{(n)} \diff{\mathbf{y}} \nonumber \\
&= \int \nabla_\theta P(\mathbf{y}|\mathbf{x}^{(n)}; \theta) R^{(n)} \diff{\mathbf{y}} \nonumber \\ 
&= \int P(\mathbf{y}|\mathbf{x}^{(n)}; \theta) \nabla_\theta \log P(\mathbf{y}|\mathbf{x}^{(n)}; \theta) R^{(n)} \diff\mathbf{y} \nonumber \\ 
&= E_{\mathbf{y}}\left[  \nabla_\theta \log P(\mathbf{y}|\mathbf{x}^{(n)}; \theta) R^{(n)} \right] \label{eq:globalreward}.
\end{align}
In Eq.~\ref{eq:globalreward}, we derived a similar equation with the gradient from the Minimum Risk Training objective \cite{shen16mrt}. However, instead of using only final reward $R^{(n)}$ and distribute it equally to every time-step, we replace the $R^{(n)}$ with the time-distributed reward $R_t^{(n)} = \sum_{i=t}^{T} \gamma^{i-t} r_i^{(n)}$ and provide more informative reward for each time-step on every sample. Therefore, we replace Eq.~\ref{eq:globalreward} to use utilize temporal structure $t=[1,..,T]$:
\begin{align}
& \nabla_\theta E_{\mathbf{y}}\left[\sum_{t=1}^{T} r_t^{(n)} | \pi_\theta \right] \nonumber \\ 
&= E_{\mathbf{y}}\left[ \sum_{t=1}^{T} r_t^{(n)} \sum_{t=1}^{T} \nabla_\theta \log P(y_t|\mathbf{y}_{<t},\mathbf{x}^{(n)}; \theta)   \right] \nonumber \\
&\approx E_{\mathbf{y}}\left[ \sum_{t=1}^{T} R_{t}^{(n)} \nabla_\theta \log P(y_t|\mathbf{y}_{<t},\mathbf{x}^{(n)}; \theta)  \right] \label{eq:localgradexpreward} \\
&\approx \frac{1}{M}\sum_{m=1}^{M} \sum_{t=1}^{T(m)} R_{t}^{(n,m)} \nabla_\theta \log P(y_t^{(n,m)}|\mathbf{y}_{<t}^{(n,m)},\mathbf{x}^{(n)}; \theta) \label{eq:localgradsample}
\end{align}
where $T$ is the length of transcription $\mathbf{y}$, $R_t^{(n)} = \sum_{i=t}^{T} \gamma^{i-t} r_i^{(n)}$ is the generalized equation for accumulated future reward based on the current state and action at time-$t$, and $\gamma$ is the discount factor to reduce the effect of future rewards. For Eq.~\ref{eq:localgradsample}, $R^{(n, m)}_t$ is the reward for the $m$-th sample based on the $n$-th utterance and time-step $t$ and $T(m)$ is the length of sample $\mathbf{y}^{(n, m)}$.
In the real world, it is impractical to integrate all possible transcription $\mathbf{y}$ to calculate the gradient of the expected reward in Eq.~\ref{eq:localgradexpreward}. Therefore, we utilize Monte Carlo sampling to sample $M$ transcription sequence $\mathbf{y}^{(n, m)} \sim P(\mathbf{y}|\mathbf{x}^{(n)}; \theta)$ from our model to calculate the gradient with empirical expectation in Eq.~\ref{eq:localgradsample}.

Since the REINFORCE gradient estimator is usually too noisy and might hinder our learning process, there are several tricks to reduce the variance \cite{greensmith2004variance, mnih2014neural}. In this paper, we normalize reward $R_t = \frac{(R_t - \mu_t)}{\sigma_t}$ where $\mu_t$ and $\sigma_t$ are the moving average and standard deviation for time-step $t$. For the final-reward $R^{(n)}$ in Eq.~\ref{eq:globalreward}, we normalize the reward across $M$ samples.

\begin{figure}[]
	\centering
	\includegraphics[width=1.0\linewidth]{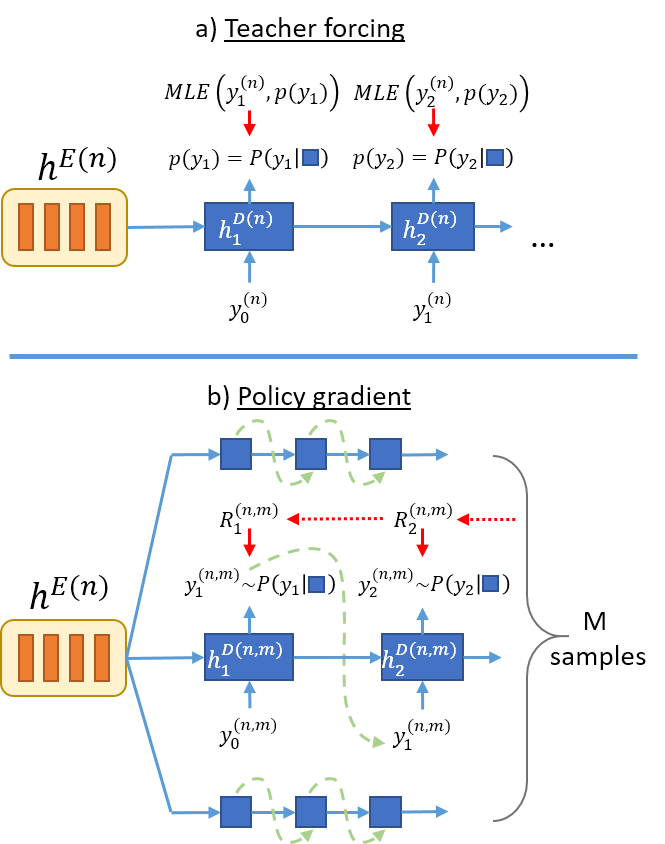}
	\caption{Comparison between teacher-forcing and policy gradient training processes. In the training stage, teacher-forcing set the model to be conditioned on the ground-truth from the dataset. Meanwhile, policy gradient method set the model to be conditioned on its own prediction from previous time-step to predicts the current time-step output probability.}
	\label{fig:teacherforcing_reinforce}
	\vspace{-0.3cm}
\end{figure}

To summarize our explanation, we provide an illustration in Fig.~\ref{fig:teacherforcing_reinforce} that compares the difference between teacher-forcing and policy gradient method for training the sequence-to-sequence model. Teacher-forcing is optimized by trying to maximize MLE objective function: 

\begin{align}
MLE(y_t^{(n)}, p(y_t)) = \sum_{c} \mathds{1}\{y_t^{(n)}=c\} * \log p(y_t=c) \label{eq:mle},
\end{align}
which is calculated per time-step based on ground-truth label $y_t^{(n)}$. In the policy gradient, first we sample $M$ sequences via Monte Carlo sampling and stop after we get an ``eos'' symbol. Then we calculate discounted reward $R_t^{(n, m)}$ for each time-step based on the future rewards.

\subsection{Reward Construction for ASR Tasks}
\vspace{-0.2cm}
\label{sec:reward}
Most ASR systems are evaluated based on edit-distance or the Levenshtein distance algorithm. Therefore, we also construct our reward function $\mathcal{R}(\mathbf{y}, \mathbf{y^{(n)}}, t)$ to calculate $r_t^{(n)}$ by utilizing the edit-distance algorithm. We define reward $r_t^{(n)}$ as
\begin{equation}
r_t^{(n)}=\begin{cases}
-(ED(\mathbf{y}_{1:t}, \mathbf{y^{(n)}})-ED(\mathbf{y}_{1:t-1}, \mathbf{y^{(n)}})) & \text{if} \quad t>1  \\
-(ED(\mathbf{y}_{1:t}, \mathbf{y^{(n)}})-|\mathbf{y^{(n)}}|) & \text{if} \quad t=1  \\
\end{cases} \nonumber
\end{equation} where $ED(\cdot, \cdot)$ is the edit-distance function between two transcriptions, $\mathbf{y}_{1:t}$ is the substring of $\mathbf{y}$ from index 1 to $t$, and $|\mathbf{y^{(n)}}|$ is the ground-truth length. Intuitively, we try to calculate whether the current new transcription at time-$t$ decreases the edit-distance compared to previous transcription, and we multiply it by -1 for a positive reward if our new edit-distance at time $t$ is smaller than the previous $t-1$ edit distance.

\section{Experiment}
\vspace{-0.2cm}
\subsection{Speech Dataset and Feature Extraction}
\vspace{-0.2cm}
In this study, we investigated the performance of our proposed method on WSJ \cite{paul92wsj} with identical definitions of training, development, and test sets as the Kaldi s5 recipe \cite{povey11asru}. We separated WSJ into two experiments using WSJ-SI84 only and WSJ-SI284 data for training. We used dev\_93 for our validation set and eval\_92 for our test set. We used the character sequence as our decoder target and followed the preprocessing steps proposed by \cite{hannun2014first}. The text from all the utterances was mapped into a 32-character set: 26 (a-z) letters of the alphabet, apostrophes, periods, dashes, space, noise, and ``eos''. In all experiments, we extracted the 40 dims + $\Delta$ + $\Delta\Delta$ (total 120 dimensions) log Mel-spectrogram features from our speech and normalized every dimension into zero mean and unit variance.

\subsection{Model Architecture}
\vspace{-0.2cm}
On the encoder side, we fed our input features into a linear layer with 512 hidden units followed by the LeakyReLU \cite{xu2015empirical} activation function. We used three bidirectional LSTMs (Bi-LSTM) for our encoder with 256 hidden units for each LSTM (total 512 hidden units for Bi-LSTM). To improve the running time and reduce the memory consumption, we used hierarchical subsampling \cite{graves2012supervised, bahdanau2016end} on the top two Bi-LSTM layers and reduced the number of encoder time-steps by a factor of 4.

On the decoder side, we used a 128-dimensional embedding matrix to transform the input graphemes into a continuous vector, followed by one-unidirectional LSTMs with 512 hidden units. For our scorer function inside the attention module, we used MLP scorers (Eq.~\ref{eq:mlpscore}) with 256 hidden units and Adam \cite{kingma2014adam} optimizer with a learning rate of $5e{-4}$.

In the training phase, we started to train our model with MLE (Eq.~\ref{eq:mle}) until convergence. After that, we continued training by adding an RL-based objective until our model stopped improving. For our RL-based objective, we tried four scenarios using different discount factors $\gamma = \{0, \, 0.5, \, 0.95\}$ and only global reward $R$ (Eq.~\ref{eq:globalreward}). To calculate the gradient based on Eq.~\ref{eq:localgradsample}, we sampled up to $M=15$ sequences for each utterance.

In the decoding phase, we extracted our transcription with a beam search strategy (beam size = 5) and normalized log-likelihood $\log P(\mathbf{Y}|\mathbf{X}; \theta)$ by dividing it by the transcription length to prevent the decoder from favoring shorter transcriptions. We did not use any language model or lexicon dictionary in this work. All of our models were implemented on the PyTorch framework\footnote{PyTorch \url{https://github.com/pytorch/pytorch/}}.

\section{Results and Discussion}
	\vspace{-0.2cm}
\begin{table}[H]
	\vspace{-0.5cm}
	\centering
	\small
	\caption{Character error rate (CER) result from baseline and proposed models on WSJ-SI84 and WSJ-SI284 datasets. All results were produced without a language model or lexicon dictionary.}
	\label{tbl:all}
	\begin{tabular}{|l|l|}
		\hline
		\multicolumn{1}{|c|}{\textbf{Models}}
		& \multicolumn{1}{c|}{\textbf{Results}} \\ \hline \hline
		\multicolumn{1}{|c|}{\textbf{WSJ-SI84}}
		& \multicolumn{1}{c|}{\textbf{CER (\%)}}          \\ \hline
		\multicolumn{2}{|c|}{\textbf{MLE}}                                                                                                                                               \\ \hline
		CTC \cite{kim2016joint}     & 20.34 \%                                       \\ \hline
		Att Enc-Dec Content \cite{kim2016joint}  & 20.06 \%                                       \\ \hline
		Att Enc-Dec Location \cite{kim2016joint}  & 17.01 \%                                       \\ \hline
		Joint CTC+Att (MTL) \cite{kim2016joint} & 14.53 \%                 \\ \hline
		Att Enc-Dec (ours) & 17.68 \%                                             \\ \hline
		\multicolumn{2}{|c|}{\textbf{MLE + RL}}                                                                                                                                               \\ \hline
		\begin{tabular}[c]{@{}l@{}}Att Enc-Dec + RL\\(final reward $R$)\end{tabular}   & \begin{tabular}[c]{@{}l@{}} 15.46 \% \end{tabular}
		\\ \hline
		\begin{tabular}[c]{@{}l@{}}Att Enc-Dec + RL\\(time reward $R_t$, $\gamma=0$)\end{tabular}   & 15.99 \%                                     \\ \hline
		\begin{tabular}[c]{@{}l@{}}Att Enc-Dec + RL\\(time reward $R_t$, $\gamma=0.5$)\end{tabular}  & 15.05 \%                                       \\ \hline
		\begin{tabular}[c]{@{}l@{}}Att Enc-Dec + RL\\(time reward $R_t$, $\gamma=0.95$)\end{tabular} & 13.90 \%     \\ \hline
		\hline		
		\multicolumn{1}{|c|}{\textbf{WSJ-SI284}}
		& \multicolumn{1}{c|}{\textbf{CER (\%)}}          \\ \hline
		\multicolumn{2}{|c|}{\textbf{MLE}} \\ \hline
		CTC \cite{kim2016joint}    & 8.97\%                                       \\ \hline
		Att Enc-Dec Content \cite{kim2016joint} & 11.08\%                                       \\ \hline
		Att Enc-Dec Location \cite{kim2016joint} & 8.17\%                                       \\ \hline
		Joint CTC+Att (MTL) \cite{kim2016joint} & 7.36\%                 \\ \hline
		Att Enc-Dec (ours)  & 7.69\%                                       \\ \hline
		\multicolumn{2}{|c|}{\textbf{MLE+RL}} \\ \hline
		\begin{tabular}[c]{@{}l@{}}Att Enc-Dec + RL\\(final reward $R$)\end{tabular}   & \begin{tabular}[c]{@{}l@{}} 7.26 \%\end{tabular}
		\\ \hline
		\begin{tabular}[c]{@{}l@{}}Att Enc-Dec + RL\\(time reward $R_t$, $\gamma=0$)\end{tabular}   &  6.64 \%                                     \\ \hline
		\begin{tabular}[c]{@{}l@{}}Att Enc-Dec + RL\\(time reward $R_t$, $\gamma=0.5$)\end{tabular}  & 6.37 \%                                       \\ \hline
		\begin{tabular}[c]{@{}l@{}}Att Enc-Dec + RL\\(time reward $R_t$, $\gamma=0.95$)\end{tabular} & 6.10 \%                                        \\
		\hline
	\end{tabular}
\end{table}

Table~\ref{tbl:all} shows all the experiment results from the WSJ-SI84 and WSJ-SI284 datasets. We compared our results with several published models such as CTC, Attention Encoder-Decoder and Joint CTC-Attention model trained with MLE objective. We also created our own baseline model with Attention Encoder-Decoder and trained only with MLE objective. The difference between our Attention Encoder-Decoder (``Att Enc-Dec (ours)'') is our decoder calculate the attention probability and context vector based on current hidden state instead of previous hidden state. We also reused the previous context vector by concatenating it with the input embedding vector. 

We explore several configurations by only using final reward $R$ and time distributed reward $R_t$ with different $\gamma = [0, 0.5, 0.95]$ values.
Our result shows that with by combining the teacher forcing with policy gradient approach improved our model performance significantly compared to a system just trained with the teacher forcing method only. Furthermore, we also found that discount factor $\gamma=0.95$ give the best performance on both datasets.

\section{Related Work}
\vspace{-0.2cm}
Reinforcement learning is a subfield of machine learning that creates an agent that interacts with its environment and learn how to maximize the rewards using some feedback signal. Many reinforcement learning applications exist, including building an agent that can learn how to play a game without any explicit knowledge \cite{mnihdqn2015, silver2016mastering}, control tasks in robotics \cite{kober2012reinforcement}, and dialogue system agents \cite{singh2000reinforcement, li2016deep}.

Not only limited to these areas, reinforcement learning has also been adopted for improving sequence-based neural network models. Ranzato et al. \cite{ranzato2015sequence} proposed an idea that combined REINFORCE with an MLE objective for training called MIXER. In the early stage of training, the first $s$ steps are trained with MLE and the remaining $T-s$ steps with REINFORCE. They decrease $s$ as the training progress over time. By using REINFORCE, they trained the model using non-differentiable task-related rewards (e.g., BLEU for machine translation). In this paper, we did not need to deal with any scheduling or mix any sampling with teacher forcing ground-truth. Furthermore, MIXER did not sample multiple sequences based on the REINFORCE Monte Carlo approximation and they were not investigate MIXER on an ASR system.

In a machine translation task, Shen et al. \cite{shen16mrt} improved the neural machine translation (NMT) model using Minimum Risk Training (MRT). Google NMT \cite{wu2016google} system combined MLE and MRT objectives to achieve better results. For ASR task, Shanon et al. \cite{shannon2017optimizing} performed WER optimization by sampling paths from the lattices used during sMBR training which might be similar to REINFORCE algorithm. But, the work was only applied on CTC-based model. From the probabilistic perspective, MRT formulation resembles the expected reward formulation used in reinforcement learning. Here, MRT formulation equally  distribute the sentence-level loss into all of the time-steps in the sample.  

In contrast, we applied the RL strategy to an ASR task and found that using final reward $R$ is not an effective method for training our system because the loss diverged and produced a worse result. Therefore, we proposed a temporal structure and applied time-distributed reward $R_t$. Our results demonstrate that we improved our performance significantly compared to the baseline system. 

\section{Conclusion}
\vspace{-0.2cm}
We introduced an alternative strategy for training sequence-to-sequence ASR models by integrating the idea from reinforcement learning. Our proposed method integrates the power of sequence-to-sequence approaches to learn the mapping between speech signal and text transcription based on the strength of reinforcement learning to optimize the model with ASR performance metric directly. We also explored several different scenarios for training with RL-based objective. Our results show that by combining RL-based objective together with MLE objective, we significantly improved our model performance compared to the model just trained with the MLE objective. The best system achieved up to 6.10\% CER in WSJ-SI284 using time-distributed reward settings and discount factor $\gamma=0.95$.

\section{Acknowledgements}
\vspace{-0.2cm}
Part of this work was supported by JSPS KAKENHI Grant Numbers JP17H06101
and JP17K00237.
\bibliographystyle{IEEEbib}
\bibliography{strings,refs}

\end{document}